\documentclass[letterpaper, 10 pt, conference]{ieeeconf} 

\IEEEoverridecommandlockouts

\let\labelindent\relax
\usepackage[inline]{enumitem}

\usepackage{cite}

\usepackage{moreverb,url}

\usepackage{verbatim}
\usepackage{amsmath}
\usepackage{graphicx}
\usepackage{amssymb}
\usepackage{array}
\usepackage[english]{babel}
\usepackage{verbatim}
\usepackage{subfigure}
\usepackage{color}
\usepackage{tabularx}
\usepackage{caption}

\usepackage{xcolor}
\pagecolor{white}

\usepackage{listings}
\usepackage[frozencache,cachedir=.]{minted} 
\usepackage[noend]{algpseudocode}
\usepackage{algorithm}
\usepackage{xpatch}

\usepackage[compact]{titlesec}
\usepackage[hidelinks]{hyperref}

\makeatletter
\xpatchcmd{\algorithmic}
  {\ALG@tlm\z@}{\leftmargin\z@\ALG@tlm\z@}
  {}{}
\makeatother

\newcommand{\aStart}[1]{
\ifmmode
\texttt{#1}_\vdash
\else 
$\texttt{#1}_\vdash$
\fi
}
\newcommand{\aEnd}[1]{
\ifmmode
\texttt{#1}_\vdash
\else 
$\texttt{#1}_\dashv$
\fi
}

\newcommand{\FriW}[1]{\emph{FriWalk}}

\begin{document}

\title{\LARGE \bf When Prolog meets generative models: a new approach \\ for managing knowledge and planning in robotic applications}

\author{Enrico Saccon, Ahmet Tikna, Davide De Martini, Edoardo Lamon, Marco Roveri, Luigi Palopoli
\thanks{This work has received funding from the European Union under NextGenerationEU (FAIR - Future AI Research - PE00000013), and from the project MUR PRIN 2020 - RIPER - Resilient AI-Based Self-Programming and Strategic Reasoning - CUP  E63C22000400001.}
\thanks{Dipartimento di Ingegneria e Scienza dell'Informazione, Università di Trento, Trento, Italy
\{enrico.saccon, ahmet.tikna, edoardo.lamon, marco.roveri, luigi.palopoli\}@unitn.it, 
davide.demartini@studenti.unitn.it}
}

\maketitle

\begin{abstract}
In this paper, we propose a robot oriented knowledge management system based on the use of the Prolog language. Our framework hinges on a special organisation of knowledge base that enables:
1. its efficient population from natural language texts using semi-automated procedures based on Large Language Models, 2. the bumpless generation of temporal parallel plans for multi-robot systems through a sequence of transformations, 3. the automated translation of the plan into an executable formalism (the behaviour trees). The framework is supported by a set of open source tools and is shown on a realistic application.
\end{abstract}



\section{Introduction}
\label{sec:intro}


We are the witnesses of a revolutionary change in robotics.  Up until
a few years back, the largest part of robots were industrial
manipulators, they operated in segregated areas, they were isolated
machines executing repetitive tasks and their level of flexibility was almost
null.

In the last few years, the rising tide of Artificial intelligence has
brought about a radical change, or at least the promise of a radical
change. The common expectation is that, in the near future, robots will be able
to operate in unstructured environments, will react "creatively" to
unanticipated events will work in large teams in which each robot
will bring its own abilities and ``expertise", and will share the same space
and interact with humans.
The bedrock of this revolution will be an effective and efficient way
to manage knowledge.
While the way humans produce and apply knowledge is only superficially understood, some
defining and commonly recognized features of human knowledge representation are:
\begin{enumerate}
\item \textbf{understandability} and \textbf{explainability}: we use different types of
  languages to accumulate information that can be shared with other
  humans;
\item \textbf{scalability}: we use hierarchies and
  conceptual links to organise massive amounts of knowledge and
  integrate new pieces as soon as they become available;
\item \textbf{usability}: conceptual entities are combined with a
  vocabulary of actions that they can generate or receive; this makes
  human able to generate action plans starting from their knowledge in
  a logically consistent and cognitively smooth fashion.
\end{enumerate}

In the past, different authors have sought ways to express knowledge
that could meet at least part of these requirements, and still be
usable in a computer program and hence in a robot.  Logic languages
like Prolog or functional languages like Lisp hold the promise to be a
\emph{lingua franca} between humans and
robots~\cite{PAULIUS201913,Cres2020}.  For instance, by using Prolog,
it is possible to express a number of relations (facts and rules) that
link existing (ground) or generic objects of our knowledge with each
other and with possible available actions. In theory, a plan can be
synthesised by applying the inference mechanism of the language.  The
language has been extended in order to support probabilistic facts and
clauses~\cite{de2007problog,Problog}, which makes it suitable to express
``imprecise knowledge".  More recently, Manaheve et
al.~\cite{manhaeve2018deepproblog} and Imanaka et al.~\cite{Neuroprolog}
have shown how to integrate Prolog with deep learning, breaking ground
for a new generation of application that integrates symbolic and 
sub-symbolic forms of artificial intelligence.
These interesting features of Prolog and logical languages mentioned
above have led to their application to construct knowledge
representations for robotics, the most famous being
KnowRob~\cite{Ten13}.  In this paper, we address two fundamental
questions.  First, while it is true that Prolog knowledge bases (\emph{KB}) are
understandable and interpretable by a human reader, they are not easy
to write for a non-specialist. On the other hand, the automatic extraction of a
strongly structured computer artefact like a Prolog knowledge base from
natural language transcript of conversations with humans could prove a
very challenging or expensive task for traditional natural language
processors. The emergence of Large Language Models~\cite{gpt4,bard}
could come to the rescue, but their abilities in this scenario are all
but tested. This
leads us to our first research question: \emph{is it possible to use
LLM to populate a robot-oriented Prolog Knowledge base} at least as
part of a semiautomated procedure?

Given that Prolog is recognised to be a very good choice for expressing
KBs in terms of logic clauses (predicates), symbolic reasoning and 
natural language processing, the second research question is whether
this programming language could be exploited to create a framework that,
starting with a LLM to process the input, is able to smoothly provide
a resilient plan and learn from possible unforeseen events.

We address the two questions above by proposing a novel Prolog-based
Knowledge Management system which has been conceived in order to:
1. simplify the population of the KB by a semi-automatic procedure
relying on the use of Large Language Models, 2. enable the seamless
generation of temporal parallel plans that support parallel actions of
multiple agents, 3. automatically translate the plan into a formalism
(the Behaviour Trees~\cite{iovino2022survey}) that can be automatically
execute by tools such as PlanSys2~\cite{martin2021plansys2}.


\section{Background}
\label{sec:bg}

\noindent\textbf{Temporal Task Planning.}
In temporal task planning, we need to reason not only about the ordering of actions in time but also about their metric duration. In the following, we will revise the elements on top of which we build our solution.
We define a \emph{(STRIPS) classical planning problem} as a tuple $CP = (F, A, I, G)$, where $F$ is the set of fluents, $A$ is a finite set of actions, $I \subseteq F$ is the initial state, and $G \subseteq F$ is a goal condition.
A literal is either a fluent or its negation.
Every action $a \in A$ is defined by two sets of literals: the preconditions, written $pre(a)$, and the effects $eff(a)$.
A \emph{classical plan} $\pi = (a_1, \cdots, a_n)$ is a sequence of actions, and it \emph{is valid} if and only if it is executable from the initial state terminating in a state that fulfills the goal $G$.
This formulation follows that presented in \cite{ghallab03}.

We define a \emph{temporal planning problem} as a tuple $TP = (F, DA, I, G)$, with $F$, $I$ and $G$ defined as in the classical planning problem, and with $DA$ being a set of \emph{durative actions}.
Following \cite{DBLP:conf/ijcai/CushingKMW07,DBLP:journals/ai/ColesFHLS09,DBLP:journals/jair/ColesCFL12}, a durative action $a \in DA$ is given by
\begin{enumerate*}[label=\roman*)]
\item two classical planning actions $a_{start}$ and $a_{end}$;
\item and a minimum $\delta_{min}(a)\in \mathbb{R}^+$ and maximum $\delta_{max}(a) \in \mathbb{R}^+$ duration, with $\delta_{min}(a) \le \delta_{max}(a)$.
\end{enumerate*}
A \emph{temporal plan} $\pi = \{tta_1, \cdots, tta_n\}$ is a set of time-triggered temporal actions, where each $tta_i$ is a tuple $\langle t_i, a_i, d_i \rangle$ where $t_i \in \mathbb{R}^+$ is the starting time, $a_i \in DA$ is a temporal action and $d_i \in [\delta_{min}(a_i), \delta_{max}(a_i)]$ is the duration of the action.
We say that $\pi$ is a \emph{valid temporal plan} if and only if it can be simulated (i.e., starting from the initial model we apply each timed triggered action $tta_i = \langle t_i, a_i, d_i \rangle \in \pi$ at time $t_i$ with duration $d_i$), and at the end of the simulation, we obtain a state fulfilling the goal condition.
For lack of space, we refer the reader to \cite{pddltwoone} for a thorough discussion on the semantics and definition of a \emph{valid temporal plan}.

State-space temporal planning is a specific approach to temporal planning. The intuition behind this approach is to combine
\begin{enumerate*}[label=\roman*)]
\item a classical forward state-space search to generate a candidate plan outline; and 
\item a temporal reasoner to check its temporal feasibility~\cite{DBLP:journals/ai/ColesFHLS09,DBLP:conf/ijcai/CushingKMW07,DBLP:journals/jair/ColesCFL12}.
\end{enumerate*}
In all these approaches, each durative action $a \in DA$ is transformed into two \emph{snap actions}: $a_{start}$ and $a_{end}$ that contain the preconditions and effects of the start and end of $a$, and additional fluents modified by $a_{start}$, $a_{inv}$ and $a_{end}$ to enforce their temporal ordering (see, e.g.,~\cite{DBLP:journals/ai/ColesFHLS09} for further details).
The resulting abstract classical problem is solved using any state-space search. 
The search extracts a classical planning and then checks if the associated temporal network is consistent, then a time-triggered plan can be computed, and the search stops since a solution has been found.
Otherwise, the search continues by computing another classical plan until either the search proves that the problem has no solution or the search bumps into a temporally consistent plan.

\noindent\textbf{Prolog.} 
Prolog is a declarative logic programming language used for the representation of knowledge and symbolic reasoning. 
In Prolog you can define a knowledge base (KB), i.e., a set of facts and rules, that can be queried to obtain information regarding the satisfiability of more complex conditions. 
In robotics, it is a great tool to represent knowledge about robots, their actions, and the environment. 
Prolog has also been used successfully in planning~\cite{prologPlanning} and has recently gained a lot of attention when paired with the natural language processing, as it can allow for easier human-robot interaction~\cite{NLPProlog1,NLPProlog2}.

We provide hereafter some intuition about the semantics of Prolog, and how it works. We refer the reader to \cite{Prolog} for a thorough description of the semantics and the peculiarities of the SWI-Prolog implementation.

In Prolog, the order in which predicates are listed matters. When queried, the interpreter will attempt to satisfy predicates in the order in which they appear in the KB. To illustrate this point, consider the following example.
Let's assume we add to the KB the two facts: \texttt{available(agent2)}, and \texttt{available(agent1)} in this order. 
When Prolog is queried with \texttt{available(X)}, it will assign \texttt{agent2} as the first value for \texttt{X} and then, when asked for another value, it will assign \texttt{agent1}, because that is the order in which they were inserted in the KB. 
Another important aspect of Prolog is that it is able to backtrack its steps when it encounters a predicate that leads to a negative result. 
%
However, to backtrack it needs to bump into a failure, and in some cases, this may not happen, and thus to avoid the search continuing forever it is important to correctly set conditions and checks.
%
%

\begin{figure*}[t]
 \includegraphics[width=1.99\columnwidth]{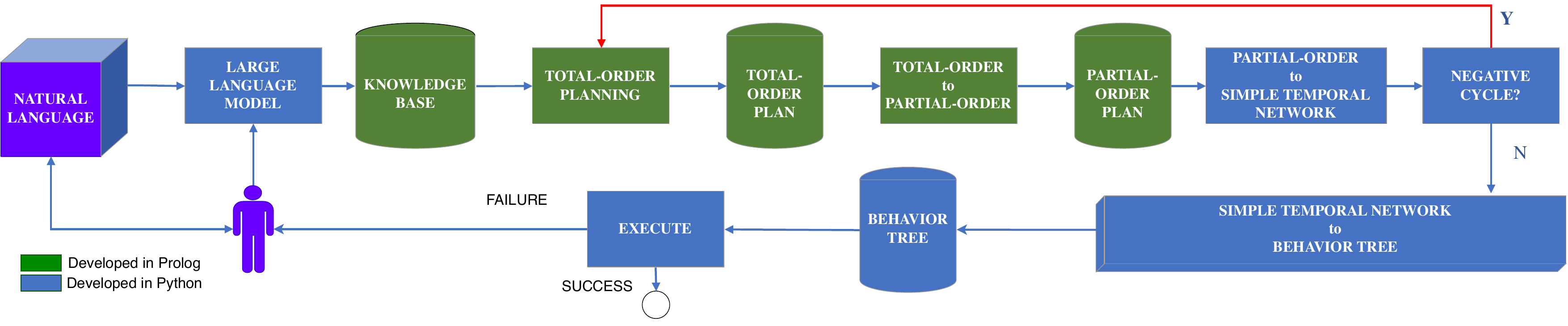}
\caption{The general diagram of the framework. The main idea is to create a feedback loop in different parts of the system to correctly modify the domain where needed and to learn from the environment.
}
    \label{fig:diagram}
    \vspace{-3mm}
\end{figure*}

\noindent\textbf{Large Language Models.}
Large Language Models (LLMs) are a type of artificial intelligence model aimed at natural language processing (NLP). Thanks to their ability to generalize and understand the context in which they are used, they have gained increasing relevance in recent years.
They are typically trained with enormous amounts of data and have hundreds of billions of parameters, which can also be fine-tuned for the task in which they have to be employed \cite{LLMsSurvey1, LLMsSurvey2}.
LLMs have been applied to a grwoing number of different fields from healthcare~\cite{LLMsHealthCare} to planning~\cite{LLMsSucc}, even though not always with the best results~\cite{LLMsCannotPlan}.
Indeed, while LLMs excel at learning complex patterns and information from vast training data, it's crucial to understand that they primarily rely on statistical associations. They do not possess genuine inferential reasoning capabilities and consequently, LLMs can struggle when confronted with tasks significantly different from the data they were trained on, no matter how extensive that may be.
Despite this, they can provide reasonably acceptable starting points for further refinements.

\section{Problem description and conceptual scheme}
\label{sec:concept}



In this section, we define the addressed problem and we describe the workflow we adopted to solve it.
We focus on the challenge of orchestrating a sequence of actions for multiple, possibly heterogeneous, agents (task planning) using the Prolog programming language. 
Our rationale for choosing Prolog is based on the fact that it is suitable for constructing knowledge bases well perceived in robotics, and it allows for symbolic reasoning.

The framework we adopted is depicted in Figure~\ref{fig:diagram}. 
We start by creating a knowledge base defined in Prolog by taking as inputs different sources. While one could directly implement the knowledge base in Prolog, the goal of the framework is to have an LLM, such as GPT or LaMBDA, that is able to parse some contents, for example, manuals, and provide the correct actions and predicates to model the problem. Having such a feature greatly increases the usability of the framework since it would be possible to specify in natural language the environment and the actions that the agents can perform. More information regarding the knowledge base will be provided in Section~\ref{sec:kb}. 
Subsequentially, we use Prolog to compute a total-order plan, considering snap actions (see Sect.~\ref{sec:bg}). The plan is obtained by leveraging a forward search approach that progresses the initial state taken from the KB applying the effects of applicable actions till the goal has been reached, exploiting the symbolic reasoning provided by Prolog. For the time being, differently from what is commonly done in temporal planning, we do not use any heuristic to guide the search (left for future work). As a result, the computed total order plan with the snap actions may be not optimal.
While constructing the total-order plan, we also save all the states in which the chosen action had its preconditions satisfied and could hence be executed. We leverage this information to build a partial order of the actions for the computed total order plan, which in turn is encoded as a Disjunctive Temporal Network, which we then strengthen to an STN by considering only the last achiever~\cite{DBLP:journals/jair/ColesCFL12} of the action precondition. 
To complete the STN, we also include constraints on the duration of each action (between the start snap action and its corresponding end~\cite{DBLP:journals/jair/ColesCFL12}). We remark that the strengthening of the DTN to the STN is not done using Prolog, but it is done in Python. For more information regarding the planning steps, see~\ref{sec:taskPlanning}.

The computed STN is then checked for consistency (by examining whether there are negative cycles in the graph representing it~\cite{STNConsistency}. If there are, then we can ask Prolog to generate a new total-order plan and subsequently a new partial-order plan and STN. If this operation fails multiple times, then there might be an error inside our knowledge base, which should be checked and corrected accordingly, hence also improving the domain. 

Once the STN is constructed and verified, we can optionally apply an optimization tool in order to shrink the STN and have actions be performed more tightly reducing the makespan or the sum of costs. Finally, the resulting STN is converted into a Behavior Tree (BT) that captures the temporal and causal dependencies among the actions in the STN, and this can be subject to execution.

While executing the constructed BT, unexpected exceptions or problems may arise because the KB used to compute the plan may be not aligned with the real world. 
If this were to happen, we need to take into consideration what led to this exception and either make the due changes to the KB, or put constraints on the agents so that the same situation won't happen again.

\section{The Knowledge base}
\label{sec:kb}
In this section, we describe the general structure of the knowledge base and how it can be populated and expanded using an LLM such as ChatGPT or BARD. 

\noindent\textbf{KB Structure.}
Prolog is a logic programming language that allows for easily creating, modifying and querying the knowledge base (See Sect. \ref{sec:bg}). 
We describe each state by a list of predicates, which define the position of the blocks and the status of the agents. For example, \texttt{ont(B, X, Y)} describes the fact that block \texttt{B} is on the table at coordinates \texttt{(X, Y)}, and \texttt{av(A)} states that agent \texttt{A} is available to be used. 
Our knowledge base is composed of predicates to describe the states, and Prolog rules that describe the actions leveraging the following structure:
\begin{minted}[fontsize=\scriptsize]{prolog}
action(Name, ValidConditions, InvalidConditions, 
       InvalidConditionsAtEnd, ConditionsOnKB, Effects).
\end{minted}
Of which this is an example:
\begin{minted}[fontsize=\scriptsize]{prolog}
action(grip_ontable_start(A, B), 
       [ontable(B, X, Y), available(A), clear(B)],
       [gripped(_, B), gripping(_, B)],
       [ontable(B, X, Y)],
       [],
       [del(available(A)), add(gripping(A, B))]).
\end{minted}

\begin{figure*}[t]
\begin{minted}[fontsize=\scriptsize]{text}
Consider the following test cases.
Each of them moves a set of boxes (b1, b2, b3, ...) from an initial state to a final state using agents(a1, a2,..).

% from b1 at the point (2,2), b2 on the table at point (1,1) to b2,b1 stacked at point (3,3).
test1 :- go( 
  [available(a1), available(a2), available(a3), ontable(b1, 2, 2), ontable(b2, 1, 1), clear(b1), clear(b2)],
  [available(a1), available(a2), available(a3), ontable(b2,3,3), on(b1, b2, 3, 3), clear(b1)]
).

% from b2,b1 stacked to b1, b2 on the table.
test2 :- go(
  [available(a1), available(a2), available(a3), ontable(b2,1,1), on(b1, b2, 1, 1), clear(b1)],
  [available(a1), available(a2), available(a3), ontable(b1,2,2), ontable(b2, 3, 3), clear(b1), clear(b2)]
).

% from b2,b1 stacked and b3 on the table to b1,b2,b3 stacked.
test3 :- go(
  [available(a1), available(a2), available(a3), ontable(b2,1,1), on(b1, b2, 1, 1), clear(b1), ontable(b3, 2, 2), clear(b3)],
  [available(a1), available(a2), available(a3), ontable(b1,3,3), on(b2, b1, 3, 3), on(b3, b2, 3, 3), clear(b3)]
).

Can you generate a prolog code containing a new test case, namely test_case, in which we use 3 agents to move the boxes
b1, b2, b3, b4 on the table, which are at (1,1),(2,2),(3,3), and (4,4), respectively, to a final stack [b1,b2,b3,b4] at 
point (5,5), which is ordered from top to bottom?
\end{minted}
\caption{Example of message used to query the LLM.}
\label{fig:LLMQuery}
\vspace{-3mm}
\end{figure*}

The variable \texttt{Name} defines the name of the action and its argument, e.g., for the action corresponding to the gripping of an agent \texttt{A} on a block \texttt{B}, we use \texttt{Name=grip(A,B)}. The four variables that follow are lists of conditions that must be checked before deciding whether to add the action or not:
\begin{itemize}
    \item \texttt{ValidConditions} contains conditions that should be verified in the current state;
    \item \texttt{InvalidConditions} contains conditions that must not be verified in the current state;
    \item \texttt{ValidConditionsAtEnd} contains conditions that must not be verified in the goal; 
    \item \texttt{ConditionOnKB} contains conditions that must be verified on the knowledge base before deciding on the action. 
\end{itemize}
The list \texttt{ValidConditionsAtEnd} checks if a condition from the goal has already been achieved and avoids actions which may make one of the contained conditions not to hold.
\texttt{ConditionOnKB} is a list used to force the Prolog interpreter to ground the variables of the action to some values. In particular, the knowledge base is composed of predicates:
\begin{itemize}
    \item \texttt{pos(X,Y)}, which indicates positions that may be used by the agents to temporarily store blocks;
    \item The predicates inside the goal state. 
\end{itemize}
The reason for adding the goal state to the knowledge base is to avoid adding trivial and useless actions. Indeed, the search in Prolog is not guided and if we were not to match the goal when choosing an action, the search would be completely unguided and the program may add useless actions such as the movement of a block in the same position. \newline
Finally, \texttt{Effects} contains a list of predicates on how to modify the current state in a new state. Each predicate is in the form of either \texttt{add(P(...))}, which adds \texttt{P(...)} to the state, or \texttt{del(P(...))}, which looks for \texttt{P(...)} in the current state and removes it.

To query for a solution, we provide the initial and final states as input parameters to the \texttt{go} function. This function serves as a convenient wrapper for the \texttt{plan} function, which is responsible for the actual plan-finding process. A detailed discussion of the this function can be found in Sect.~\ref{sec:taskPlanning}. An example query is the following one: 
\begin{minted}[fontsize=\scriptsize]{prolog}
test1 :- go([available(a1), ontable(b1, 1, 1), clear(b1)],
            [available(a1), ontable(b1, 2, 2), clear(b1)]).
\end{minted}

\noindent\textbf{LLMs.}
Large Language Models are a sort of machine learning model developed to comprehend and generate human language.
They are often built upon transformer~\cite{Transformer} networks, which utilize self-attention mechanisms to gain a better understanding of the context of the words in a sentence.
Another innovation brought by transformer networks is positional encoding, which allows the model to process words in the sentence in parallel, rather than sequentially while providing the model with information about the positions of the words in a sentence.
This substantially improves the effectiveness and speed of the model.
The ability of large language models to develop broad knowledge from massive datasets and to infer semantic relationships between textual entities enables them to solve a variety of natural language processing problems.
In our approach, LLMs are employed in order to construct the representation of the environment and generate initial and goal states in Prolog format for given problems specified as natural language queries.
LLMs are provided with queries involving multiple test cases described above.
In the query configuration, there are multiple examples of test cases with explanatory comments and a problem description, as shown in Figure \ref{fig:LLMQuery}.






In response to the same query, LLMs can generate different outputs. 
We set the temperature parameter to zero so that LLMs operate deterministically.


\section{Task Planning}
\label{sec:taskPlanning}

In this section, we first describe how we compute a total order plan to solve the task planning problem. We then discuss how we extract a partial-order plan from the total-order one, followed by its transformation into an STN. Finally we provide a sketch of how we obtain behaviour trees from the STN. 

\begin{table*}[t]
    \fontsize{7pt}{7pt}\selectfont
    \centering
    \begin{tabular}{lrl}
\multicolumn{1}{c}{State}  & \multicolumn{2}{c}{Action} \\
\texttt{[av($a_1$), av($a_2$), ont($b_1$, 1, 1), ont($b_2$, 2, 2), clr($b_1$), clr($b_2$)]}               & $a_1:$ & \aStart{grip($a_1$, $b_2$)} \\
\texttt{[gripping($a_1$, $b_2$), av($a_2$), ont($b_1$, 1, 1), ont$(b_2$, 2, 2), clr($b_1$), clr($b_2$)]}  & $a_2:$ & \aEnd{grip($a_1$, $b_2$)} \\
\texttt{[gripped($a_1$, $b_2$), av($a_2$), ont($b_1$, 1, 1), ont($b_2$, 2, 2), clr($b_1$)]}               & $a_3:$ & \aStart{move\_block($a_1$, $b_2$, 2, 2, 3, 3)} \\
\texttt{[moving($a_1$, $b_2$, 2, 2, 3, 3), av($a_2$), ont($b_1$, 1, 1), clr($b_1$)]}                      & $a_4:$ & \aStart{grip($a_2$, $b_1$)} \\
\texttt{[gripping($a_2$, $b_1$), moving($a_1$, $b_2$, 2, 2, 3, 3), clr($b_1$)]}                           & $a_5:$ & \aEnd{grip($a_2$, $b_1$)} \\
\texttt{[gripped($a_2$, $b_1$), moving($a_1$, $b_2$, 2, 2, 3, 3)]}                                        & $a_6:$ & \aEnd{move\_block($a_2$, $b_1$, 1, 1, 3, 3)} \\
\texttt{[gripped($a_2$, $b_1$), ont($b_2$, 3, 3), clr($b_2$), av($a_1$)]}                                 & $a_7:$ & \aStart{move\_block($a_2$, $b_1$, 1, 1, 3, 3)} \\
\texttt{[moving($a_2$, $b_1$, 1, 1, 3, 3), ont($b_2$, 3, 3), clr($b_2$), av($a_1$)]}                      & $a_8:$ & \aEnd{move\_block($a_2$, $b_1$, 2, 2, 3, 3)} \\
\texttt{[on($b_1$, $b_2$, 3, 3), clr($b_1$), av($a_2$), ont($b_2$, 3, 3), clr($b_2$), av($a_1$)]} &   \\
    \end{tabular}
    \caption{Table describing how the actions change the prior state. \texttt{av($a_i$)} states that agent $a_i$ is available, \texttt{ont($b_i$, X, Y)} states that block $b_i$ is in position \texttt{(X,Y)}, \texttt{on($b_i$, $b_j$, X, Y)} states that block $b_i$ is on top of block $b_j$ in position \texttt{(X,Y)}.}.
    \label{tab:fromTotToPart}
    \vspace{-3mm}
\end{table*}

\subsection{Total-Order Plan}
To compute a total-order plan we have developed the \texttt{plan} function (see Algorithm~\ref{algo:algo}). This function recursively checks whether an action from the available action list can be scheduled for execution, by:
\begin{enumerate*}[label=\roman*)]
    \item verifying whether the action's preconditions hold in the current state; and
    \item assessing that none of the undesired predicates holds in the current state or in the final state. 
\end{enumerate*}
If both conditions are met, the function applies the effects specified by the action, resulting in a new state. Subsequently, the function is invoked recursively to determine whether the current state matches the goal state. If the two states match, the process terminates and the computed plan is returned, otherwise, it continues to search for a viable plan. 

The function also checks that the depth of the recursion (length of the total order plan) does not exceed a given threshold by forcing a fail, which in turns triggers a backtrack to search for other solutions. The depth has been limited to compensate for the uninformed search performed, which may lead to very deep, still valid searches, that may not lead anywhere.

\subsection{Partial-Order Plan and STN}

To obtain the partial order plan we compare the preconditions of each newly chosen action with the effects of the previous actions. Indeed, while the chosen action was correctly added at a given moment since its preconditions were verified, this does not capture all the causality relationships between the different actions. What we want to capture are all the \textit{achievers}, that is, actions whose effects allow the last added action to be executed. The goal of this step is to obtain a graph of temporal-causal relationships between the different actions so that an action can be executed only when and as soon as its preconditions are satisfied. 

To correctly bind actions between each other, we split the actions into a starting and a terminating action, e.g., \texttt{move\_block} becomes \aStart{move\_block} (start) and \aEnd{move\_block} (end). In this way, we can safely state that another action can start only when the previous one is finished, e.g., a grip on a block can only start when the move on the same block has ended. Moreover, constraints on the duration of the actions have been put in place, that is, a terminating action cannot happen after a certain amount of time $d_a$ from the starting of the action: $\aEnd{a}-\aStart{a}\leq d_a$. 
Creating such a graph allows for multiple actions to be carried out at the same time. 

We create this graph by calling the \texttt{partial\_order} predicate each time an action is added to the total order plan. Such predicate takes the considered action and the list of previous actions and recursively checks if any of the preconditions of the chosen action are satisfied by the effects of another previous action. If they are, then there is a causal link between the two actions. The actions that do not have a causality relationship can be executed in parallel. 

\begin{figure}[htp]
    \centering
     \includegraphics[width=0.8\linewidth]{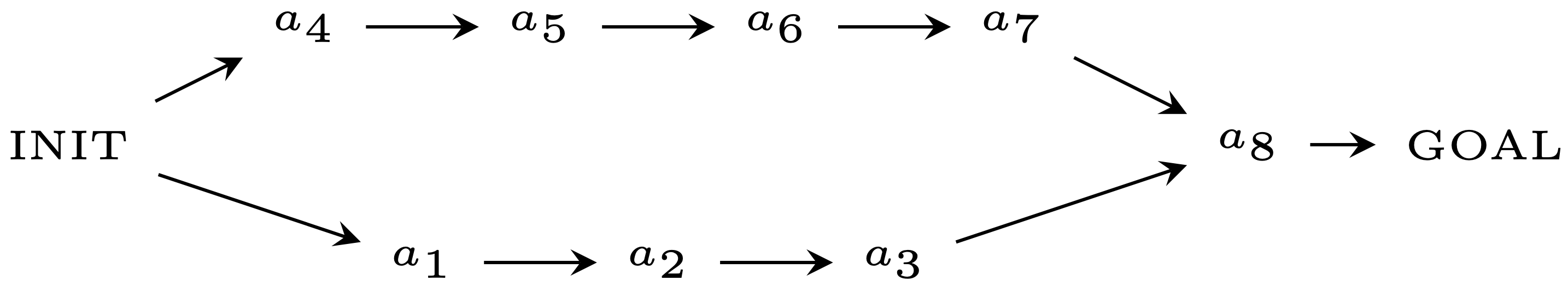}
    \caption{Graph representing the partial-order plan obtained from the total-order plan described in Table~\ref{tab:fromTotToPart}}
    \label{fig:PartialOrderGraph}
    \vspace{-3mm}
\end{figure}

Consider the total-order plan shown in Table~\ref{tab:fromTotToPart}. By applying the above function we obtain, for each action, a list of achievers that are needed for the pre-conditions of the action (see Table \ref{tab:fromTotToPartAchievers}).
\begin{table}[htp]
\centering
    \begin{tabular}{crl}
        \multicolumn{2}{c}{Action} & \multicolumn{1}{c}{Achivers} \\
            \hline
        $a_1$ & {\footnotesize\aStart{grip(a1,b2)}}                & [$a_0$] \\
        $a_2$ & {\footnotesize\aEnd{grip(a1,b2)}}                  & [$a_1$] \\
        $a_3$ & {\footnotesize\aStart{move\_block(a1,b2,2,2,3,3)}} & [$a_2$] \\
        $a_4$ & {\footnotesize\aStart{grip(a2,b1)}}                & [$a_0$] \\
        $a_5$ & {\footnotesize\aEnd{grip(a2,b1)}}                  & [$a_4$] \\
        $a_6$ & {\footnotesize\aStart{move\_block(a2,b1,1,1,3,3)}} & [$a_5$] \\
        $a_7$ & {\footnotesize\aEnd{move\_block(a2,b1,1,1,3,3)}}   & [$a_6$] \\
        $a_8$ & {\footnotesize\aEnd{move\_block(a1,b2,2,2,3,3)}}   & [$a_7$, $a_3$] \\
    \end{tabular}
    \caption{List of achievers for the total order plan in Table \ref{tab:fromTotToPart}.}
    \label{tab:fromTotToPartAchievers}
    \vspace{-3mm}
\end{table}
The construction of the graph from this list leads to a graph similar to the one in Figure~\ref{fig:PartialOrderGraph}, from which we can see that the two series of action gripping and moving the blocks can be run in parallel for the majority of the time. 

At this point, we want to obtain an STN from the partial-order. To do this, for each action, we keep only the earliest timestamp at which the action was executable. Moreover, we enforce the constraints on the duration of the action by inserting backward links of negative weight between the nodes. 
Once the STN has been built, we check that it is consistent, i.e., that there are no negative cycles. The possibility of finding negative cycles comes from the fact that we consider a lower and an upper bound for the action duration, which increases the resilience of the final plan to possible delays in the real-world execution of the actions. So the constraint on the duration $\aEnd{a}-\aStart{a}\leq d_a$ becomes $l_a\leq\aEnd{a}-\aStart{a}\leq u_a$, where $l_a$ and $u_a$ are the lower and upper bounds on the duration of action $a$, respectively. \newline
The final phase, involving the construction and validation of the STN, was not implemented in Prolog, since we exploited the Networkx Python framework (\url{https://networkx.org/}).

\begin{algorithm}
\scalebox{0.8}{
\begin{minipage}{1.2\linewidth}
\begin{algorithmic}
\Function{plan}{$State$, $Goal$, $Been\_list$, $Actions$, $Times$, $MaxDepth$}
    \If {$State = Goal$}
        \State $print(Actions)$
    \Else
        \State $length(Actions) < MaxDepth$
        \State $choose\_actions(Name, Preconditions, Effects)$
        \State $check\_conditions(Preconditions, State)$
        \State $Child\_state \gets change\_state(State,Effects)$
        \If {$Child\_state \not\in Been\_list$}
            \State $Stack(Child\_state, Been\_list)$
            \State $Stack(Name, Actions)$
            \State $partial\_order(Preconditions, Been\_list, Time)$
            \State recursively call $plan$
        \EndIf
    \EndIf
\EndFunction

\Function{partial\_order}{$Conditions$, $Action\_list$, $Achievers$}
    \For {$Action_i \in Action\_list$}
        \If {$achiever(Conditions, Action_i)$}
            \State $Achievers \cup \{i\}$
        \EndIf
    \EndFor
\EndFunction
\end{algorithmic}
\end{minipage}
}
\caption{The pseudo-code for the total-order plan and for the partial-order plan.}
\label{algo:algo}
\end{algorithm}

\subsection{Behaviour Trees}
The final step we implemented of the diagram shown in Figure~\ref{fig:diagram} is the modelling of behaviour trees from the STN. We will not delve into deep in the description of this step as a much more complete report is available~\cite{BTReport}. 

Starting from the root of the STN, we compute a deep-first search of the network adding a sequential behaviour sub-tree each time that an action has only one exiting link, or a parallel behaviour sub-tree when the action has multiple exiting links. A parse action is done before calling the function to create the BT in order to remove the backward links. This is done to create narrower behaviour trees reducing the number of waiting actions that would have otherwise been inserted. 

Every time a start action is met during the DFS, a sub-tree is created which is in charge of starting the action, that is, checking the preconditions and applying the effects at the start. Similarly, every time an end action is met, the algorithm inserts a sub-tree, whose role is to check that the conditions to terminate the action are present and to apply the correct effects at the end. 

\section{An example application}
\label{sec:example}

Here we discuss an example of the application of the proposed framework. We first show the results obtained by using different LLMs to update the knowledge base with new tests, and then we describe the simulation test environment.

\subsection{LLMs Result}

\begin{table}[t!]
    \centering\scalebox{0.84}{
    \renewcommand{\arraystretch}{1.5}
    \begin{tabular}{@{}p{1.0cm}|p{1.35cm}|p{1.3cm}|p{1.85cm}|p{1.45cm}|p{0.75cm}@{}}
	    ~ & \# of predicates (avg)
        & \# of literals \newline (avg)  
        & \# of error in predicates (avg)
        & \# of error in \newline literals (avg)
        & success rate
        \\\hline 
        BARD
        & 16.5 & 33.4 & 2.4 & 5.7 & 0.4 \\ \hline
        GPT-3.5
        & 16.5 & 33.4 & 1.3 & 3 & 0.4 \\  \hline
        GPT-4  
        & 16.5 & 33.4 & 0.6 & 1.6 & 0.8
    \end{tabular}}
    \caption{Large Language Model Evaluation}
	\label{tab:confPar}
 \vspace{-3mm}
\end{table}

We evaluate the performance of 3 LLMs, namely GPT-3.5, GPT-4, and BARD.
We designed 10 different scenarios, where the framework was assigned the task of picking and placing a number of boxes ranging from 3 to 5, with a number of manipulators varying from 2 to 4.
In the experiments, the LLMs operate under identical configurations. 
As depicted in Table-\ref{tab:confPar}, GPT-4 commits the least number of errors.
The most frequent mistake made by LLMs is stacking boxes in the wrong order at the correct coordinates.
The majority of the mistakes occur in the final states.

\subsection{Environment and Technologies Used in the Simulation}
\begin{figure}[htp]
    \centering
    \includegraphics[width=0.5\columnwidth]{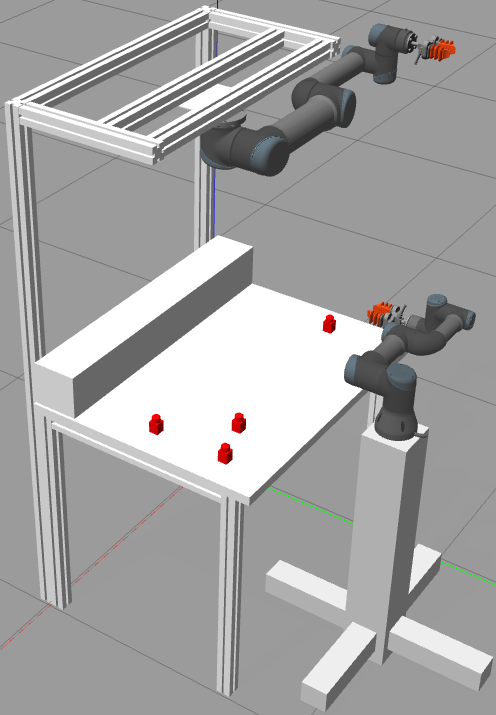}
    \caption{Gazebo simulation environment. A video of the experiments is available in the multimedia extension and in \url{https://youtu.be/zNF1T8efGW0}.}
    \label{fig:gazebosim}
    \vspace{-3mm}
\end{figure}

We developed a validation scenario leveraging ROS and Gazebo (well-known frameworks in the robotic community). 
For our validation, we considered two different robots, a UR5e and a UR3e from Universal Robots. In our specific scenario, we adjusted the UR5 upside down attached to a workbench, while the UR3 is mounted on its custom base and faces the workbench, as shown in Figure~\ref{fig:gazebosim}. In order to pick and place objects both robots mount a SoftRobotics two finger gripper. \newline
In the considered scenario the manipulators have to perform some pick and place tasks to arrange a set of mega blocks in a defined positions. 
We did not use any ROS package neither for planning nor for executing the behaviour trees. Instead, we implemented our version of the behaviour trees and to actually move the robots we sent messages to the correct topics.  


\section{Conclusions and future work}
\label{sec:conclusions}
In this paper, we have shown a robot-oriented knowledge management
system based on Prolog.  A key feature of the approach is its effective
integration with large language models, which simplifies the
generation of the knowledge base from an informal textual description,
and a bump-less procedure that produces an executable plan to
orchestrate the parallel operations of a set of robotic agents.

Many issues remain open and will require future investigations. The
most important research directions that we intend to follow are:
1. improving the use of LLMs with the automated correction of menial
errors and the detection of logical inconsistency that could reveal
that the LLM is "hallucinating", 2. the integration of probabilistic
clauses that can be associated with uncertain events or perceptions
(e.g., "This could be a hammer with 0.65 probability"), 3. dynamic
generation of new clauses and facts when the system comes across an
unmodelled aspect of its operation domain (e.g., object too heavy to be
lifted by one arm), 4. optimization of the STN or of the BT focusing 
on the temporal span, flexibility and resilience, 5. improving the
search for the total plan guiding it with an heuristic.




\bibliographystyle{IEEEtran}
\bibliography{reference}

\end{document}